\documentclass[twoside,11pt]{article}

\usepackage{multirow}
\usepackage{graphicx}
\usepackage{float}
\usepackage[table]{xcolor} 
\usepackage{booktabs}
\usepackage{tablefootnote}
\definecolor{Gray}{gray}{0.9}
\newcommand{\gray}{\rowcolor{Gray}}
\usepackage{caption}
\definecolor{Yellow}{rgb}{0.9,0.9,0}

\usepackage{todonotes}

\usepackage[preprint]{jmlr2e}

\usepackage{lastpage}

\ShortHeadings{AdvSecureNet: A Python Toolkit for Adversarial Machine Learning}{Catal and Günther}
\firstpageno{1}

\begin{document}

\title{AdvSecureNet: A Python Toolkit for Adversarial Machine Learning}

\author{\name Melih Catal \email melihcatal@gmail.com \\
       \addr Software Evolution and Architecture Lab \\
       University of Zurich, Switzerland
       \AND
       \name Manuel Günther \email guenther@ifi.uzh.ch \\
       \addr Artificial Intelligence and Machine Learning Group\\
       University of Zurich, Switzerland
}

\editor{My editor}

\maketitle

\begin{abstract}
Machine learning models are vulnerable to adversarial attacks. Several tools have been developed to research these vulnerabilities, but they often lack comprehensive features and flexibility. We introduce AdvSecureNet, a PyTorch based toolkit for adversarial machine learning that is the first to natively support multi-GPU setups for attacks, defenses, and evaluation.  It is the first toolkit that supports both CLI and API interfaces and external YAML configuration files to enhance versatility and reproducibility.  The toolkit includes multiple attacks, defenses and evaluation metrics. Rigiorous software engineering practices are followed to ensure high code quality and maintainability. The project is available as an open-source project on GitHub at \url{https://github.com/melihcatal/advsecurenet} and installable via PyPI.
\end{abstract}


\section{Introduction}

Machine learning models are widely used in fields such as self-driving cars \citep{bojarski2016end}, facial recognition \citep{parmar2014face,guenther2016survey}, and medical imaging \citep{mintz2019introduction}, as well as in natural language processing tasks like chatbots \citep{GPT3} and translation services \citep{popel2020transforming}. However, these models are vulnerable to adversarial attacks -- subtle input modifications that can deceive the models \citep{fgsm_goodfellow, Szegedy2013IntriguingPO}, which can compromise their integrity, confidentiality, or availability \citep{khalid2021exploiting_attacks}.

Several libraries, such as ART \citep{art2018}, AdverTorch \citep{ding2019advertorch}, and CleverHans \citep{cleverhans}, have been developed to research these vulnerabilities by providing tools for implementing attacks, defenses, and evaluation metrics. However, these libraries often lack key features necessary for comprehensive research and experimentation, such as native multi-GPU support, integrated CLI and API interfaces, and support for external configuration files.

To address these limitations, we introduce \textbf{AdvSecureNet} (Adversarial Secure Networks), a comprehensive and flexible Python toolkit that supports multiple adversarial attacks, defenses, and evaluation metrics, optimized for multi-GPU setups. It includes a command-line interface (CLI) and an application programming interface (API), providing users with versatile options for experimentation and research. This paper outlines the features, design, and contributions of AdvSecureNet to the adversarial machine learning community.


\begin{table}
\centering
\resizebox{\textwidth}{!}{%
\begin{tabular}{>{\bfseries}lccccccc}
\toprule
Feature & \textbf{AdvSecureNet} & \textbf{IBM Art} & \textbf{AdverTorch} & \textbf{SecML} & \textbf{FoolBox} & \textbf{Ares}  & \textbf{CleverHans} \\
\midrule
\gray Actively Maintained & \checkmark & \checkmark & $\times$ & $\times$ & $\times$ & $\times$ & $\times$ \\
Last Year of Contribution  & 2024 & 2024 & 2022 & 2024 & 2024 & 2023 & 2023 \\
\gray Pytorch Support & \checkmark & \checkmark & \checkmark & \checkmark & \checkmark & \checkmark & \checkmark  \\
 Tensorflow Support & $\times$ & \checkmark & $\times$ & \checkmark & \checkmark & $\times$ & \checkmark \\
\gray Number of Adversarial Attacks & 8 & 60 & 17 & 39\footnotemark[1] & 31 & 28 & 8 \\
Number of Defenses & 2 & 37 & 3 & - & - & 3 & 1 \\
\gray  Number of Evaluation Metrics & 6 & 5 & - & -  & 2 & 1 & 2  \\
Integrated Multi-GPU Support & \checkmark & $\times$ & $\times$ & $\times$ & $\times$ & Limited\footnotemark[2] & $\times$ \\
\gray API Usage & \checkmark & \checkmark & \checkmark & \checkmark & \checkmark & \checkmark & \checkmark \\
CLI Usage & \checkmark & $\times$ & $\times$ & $\times$ & $\times$ & Limited\footnotemark[2] & $\times$ \\
\gray External Config File & \checkmark & $\times$ & $\times$ & $\times$ & $\times$ & Limited\footnotemark[2] & $\times$ \\
 GH Stars & 2 & 4.6k & 1.3k & 138 & 2.7k & 468 & 6.1k \\
\gray GH Forks & 0 & 1.1k & 193 & 23 & 422 & 88 & 1.4k \\
 Number of Contributors & 1 & 105 & 17 & 8 & 32 & 6 & 110 \\
\gray Number of Citations & - & 571 & 222 & 14 & 677 & 291 & 400 \\
\bottomrule
\end{tabular}
}
\caption{Feature Comparison of AdvSecureNet vs. Existing Libraries (26.06.2024)}
\label{table:library_comparison}
\end{table}

\footnotetext[1]{SecML supports attacks from CleverHans \citep{cleverhans} and FoolBox \citep{rauber2017foolbox}.}
\footnotetext[2]{This feature is only available for adversarial training.}

\section{AdvSecureNet Features}

\textbf{Adversarial Attacks and Defenses:} AdvSecureNet supports a diverse range of evasion attacks on computer vision tasks, including gradient-based, decision-based, single-step, iterative, white-box, black-box, targeted, and untargeted attacks \citep{khalid2021exploiting_attacks}. AdvSecureNet also includes defense mechanisms such as adversarial training \citep{fgsm_goodfellow, Kurakin2016AdversarialEI}, which incorporates adversarial examples into the training process to enhance model resilience, and ensemble adversarial training \citep{Tramr2017EnsembleAT}, which leverages multiple models or attacks to develop a more resilient defense strategy.

\textbf{Evaluation Metrics:} AdvSecureNet supports metrics like accuracy, robustness, transferability, and similarity. Accuracy measures performance on benign data, robustness assesses resistance to attacks, transferability evaluates how well adversarial examples deceive different models, and similarity quantifies perceptual differences using PSNR \citep{hore2010image} and SSIM \citep{Wang2004ImageQA}.

\textbf{Multi-GPU Support:} AdvSecureNet is optimized for multi-GPU setup, enhancing the efficiency of training, evaluation, and adversarial attack generation, especially for large models and datasets. This parallel GPU utilization aims to reduce computational time, making the toolkit ideal for large-scale experiments.

\textbf{Interfaces and Configuration:} AdvSecureNet offers both CLI and API interfaces. The CLI allows for quick execution of attacks, defenses, and evaluations, while the API provides advanced integration and extension within user applications. The toolkit also supports YAML configuration files for easy parameter tuning and experimentation, enabling users to share experiments, reproduce results, and manage setups effectively.

\textbf{Built-in Models, Datasets and Target Generation:} AdvSecureNet supports all PyTorch vision library models and well-known datasets like CIFAR-10, CIFAR-100, MNIST, FashionMNIST, SVHN, and ImageNet, allowing users to start without additional setup. Additionally, it can automatically generate adversarial targets for targeted attacks to simplify the attack configuration process. Users can still provide target labels manually and use custom datasets and models if desired.

\section{Design and Implementation}


AdvSecureNet is a modular, extensible, and user-friendly toolkit built on PyTorch for efficient computation and GPU acceleration. It includes core modules for attacks, defenses, evaluation metrics, and utilities, each with well-defined interfaces. The toolkit follows best practices in software engineering, featuring comprehensive testing, documentation, and CI/CD pipelines. It adheres to PEP 8 guidelines and uses Black for code formatting, along with tools like Pylint \citep{pylint} and MyPy \citep{mypy} for static code analysis and type checking. SonarQube \citep{sonarqube2024} and Radon \citep{radon} provide insights into code quality and complexity. The project is hosted on GitHub under MIT license. Documentation is available on GitHub Pages, which includes detailed guidance on installation, usage, and comprehensive API references. AdvSecureNet is also available as a pip package on PyPI for easy installation and use across various environments.

\begin{table}
\centering
\resizebox{\textwidth}{!}{%
\begin{tabular}{>{\bfseries}lccccc}
\toprule
Metric & Toolkit & Dataset & Single GPU Time (min) & Multi-GPU Time (min) & Speedup \\
\midrule
\multirow{6}{*}{FGSM Attack} 
                             & AdvSecureNet & CIFAR-10 & 0.4 & 0.37 (4 GPUs), \textbf{0.24 (7 GPUs)} & 1.09x (4 GPUs), 1.64x (7 GPUs) \\
                             & IBM ART & CIFAR-10 & 0.82 & N/A & N/A \\
                             & CleverHans & CIFAR-10 & 0.25 & N/A & N/A \\
                             & ARES & CIFAR-10 & 0.45 & N/A & N/A \\
                             & FoolBox & CIFAR-10 & 0.38 & N/A & N/A \\
                             & AdverTorch & CIFAR-10 & \textbf{0.19} & N/A & N/A \\
\midrule
\multirow{6}{*}{PGD-20 Attack}  
                             & AdvSecureNet & CIFAR-10 & 3.48 & 2.47 (4 GPUs), \textbf{1.78 (7 GPUs)} & 1.41x (4 GPUs), 1.95x (7 GPUs) \\
                             & IBM ART & CIFAR-10 & 11.0 & N/A & N/A \\
                             & CleverHans & CIFAR-10 & 3.87 & N/A & N/A \\
                             & ARES & CIFAR-10 & \textbf{3.05} & N/A & N/A \\
                             & FoolBox & CIFAR-10 & 3.67 & N/A & N/A \\
                             & AdverTorch & CIFAR-10 & 3.63 & N/A & N/A \\
\midrule
\multirow{3}{*}{Adversarial Training on CIFAR-10} 
                             & AdvSecureNet & CIFAR-10 & 5.07 & 4.03 (4 GPUs), \textbf{2.77 (7 GPUs)} & 1.26x (4 GPUs), 1.83x (7 GPUs) \\
                             & ARES & CIFAR-10 & 15.9 & 12.0 (4 GPUs), 12.8 (7 GPUs) & 1.33x (4 GPUs), 1.24x (7 GPUs) \\
                             & IBM ART & CIFAR-10 & \textbf{4.87} & N/A & N/A \\
\midrule
\multirow{3}{*}{Adversarial Training on ImageNet} 
                             & AdvSecureNet & ImageNet & \textbf{240} & 33 (4 GPUs), \textbf{30 (7 GPUs)} & 7.27x (4 GPUs), 8x (7 GPUs) \\
                             & ARES & ImageNet & 627 & 313 (4 GPUs), 217 (7 GPUs) & 2.0x (4 GPUs), 2.89x (7 GPUs) \\
                             & IBM ART & ImageNet & 323 & N/A & N/A \\
\bottomrule
\end{tabular}
}
\caption{\small \textbf{Performance Benchmark for AdvSecureNet and Other Toolkits.} Training times represent one epoch, and attack times represent the duration needed to run over the training dataset. Evaluations were conducted using ResNet-50 with Python 3.10.9 and 8x GeForce RTX™ 2080 Ti Turbo 11G GPUs. Code: \url{https://github.com/melihcatal/advsecurenet_benchmark}.}
    \label{tab:performance_benchmark}
\label{tab:performance_benchmark}
\end{table}

\section{Related Work and Comparison with Existing Toolkits}

The burgeoning field of machine learning security has led to the development of several libraries designed to aid researchers. Notable among these are ART \citep{art2018}, AdverTorch \citep{ding2019advertorch}, SecML \citep{melis2019secml}, FoolBox \citep{rauber2017foolbox}, Ares \citep{dong2020benchmarkingares}, and CleverHans \citep{cleverhans}. ART, developed by IBM, is recognized for its extensive range of attacks, defenses, and support for multiple frameworks. AdverTorch, created by Borealis AI, focuses on PyTorch and offers a wide array of attacks, though it lacks support for adversarial training. CleverHans, one of the earliest libraries in the field, was initially designed for testing adversarial attacks, and as a result, has limited defensive capabilities. SecML and Ares, while smaller in scale, provide unique features; Ares, for instance, supports distributed training and external configuration files. FoolBox is distinguished by its diverse attack portfolio and support for multiple frameworks, but it does not offer defensive methods. Unfortunately, many of these libraries are no longer maintained. Table \ref{table:library_comparison} provides a detailed comparison of the features offered by these libraries.

AdvSecureNet stands out among existing adversarial machine learning toolkits in both its features and performance. Regarding features, AdvSecureNet is one of the few toolkits that are actively maintained, which is crucial for ongoing support. While IBM ART offers the most extensive attacks and defenses, AdvSecureNet provides a balanced selection, including adversarial and ensemble adversarial training for defense and a diverse range of attacks for evasion. AdvSecureNet distinguishes itself by being the first toolkit to natively support multi-GPU setups for adversarial attacks, defenses, and evaluation, whereas ARES only supports distributed adversarial training. This makes AdvSecureNet ideal for large-scale experiments. It is also the first toolkit that fully supports both CLI and API usages and external YAML configuration files, aiding researchers in sharing and reproducing experiments.

AdvSecureNet shows its strength in performance, achieving faster execution times on multi-GPU setups compared to other toolkits. As shown in Table \ref{tab:performance_benchmark}, AdvSecureNet’s multi-GPU PGD attack time (1.78 minutes) outperforms ARES’s best single GPU time (3.05 minutes). In adversarial training on CIFAR-10, AdvSecureNet reduces training time from 5.07 minutes on a single GPU to 2.77 minutes with 7 GPUs, a speedup of 1.83x. AdvSecureNet's performance is even more impressive on ImageNet, reducing training time from 240 minutes on a single GPU to 30 minutes with 7 GPUs, which is an 8x speedup. In comparison, ARES reduces training time from 627 minutes on a single GPU to 217 minutes with 7 GPUs, a less efficient speedup of 2.89x. IBM ART, which does not natively support multi-GPU setups, remains at 323 minutes on a single GPU. The results show that AdvSecureNet provides superior performance and scalability, making it an ideal choice for large-scale adversarial machine learning experiments.


\section{Future Work and Conclusion}
The AdvSecureNet toolkit is an ongoing project, and we plan to continue improving and expanding its capabilities. Currently, the toolkit focuses on evasion attacks and defenses in computer vision tasks, but we aim to extend its functionality to other domains, such as natural language processing. Additionally, we plan to incorporate other aspects of the trustworthiness of machine learning models, including fairness and interpretability. 

In conclusion, AdvSecureNet is a comprehensive toolkit for adversarial machine learning research, offering a wide range of attacks, defenses, datasets, and evaluation metrics in addition to multi-GPU support, CLI and API interfaces, as well as external configuration files. By providing a flexible and efficient platform for experimentation, AdvSecureNet aims to advance the field of adversarial machine learning.

\newpage
\bibliography{references}

\end{document}